\documentclass[11pt,a4paper]{article}

\usepackage[utf8]{inputenc}
\usepackage[T1]{fontenc}
\usepackage{amsmath,amssymb}
\usepackage{graphicx}
\usepackage{booktabs}
\usepackage{hyperref}
\usepackage{natbib}
\usepackage{microtype}
\usepackage[margin=1in]{geometry}

\title{Continuum Memory Architectures for Long-Horizon LLM Agents}
\author{Joe Logan\\
\textit{Mode7 GK}\\
Tokyo, Japan\\
\texttt{hello@joelogan.co}}
\date{}

\begin{document}

\maketitle

\begin{abstract}
Retrieval-augmented generation (RAG) has become the default strategy for providing large language model (LLM) agents with contextual knowledge. Yet RAG treats memory as a stateless lookup table: information persists indefinitely, retrieval is read-only, and temporal continuity is absent. We define the \textit{Continuum Memory Architecture} (CMA), a class of systems that maintain and update internal state across interactions through persistent storage, selective retention, associative routing, temporal chaining, and consolidation into higher-order abstractions. Rather than disclosing implementation specifics, we specify the architectural requirements CMA imposes and show consistent behavioral advantages on tasks that expose RAG's structural inability to accumulate, mutate, or disambiguate memory. Preliminary evaluation across four behavioral probes (knowledge updates, temporal association, associative recall, contextual disambiguation) provides initial evidence that CMA-class behaviors yield advantages on tasks that stress memory dynamics, while highlighting open challenges around latency, drift, and interpretability.
\end{abstract}

\section{Introduction}

LLM-based agents are increasingly expected to operate across extended time horizons: remembering user preferences, tracking evolving projects, or coordinating long-running workflows. The prevailing approach equips agents with a vector database and retrieves the top-$k$ semantically similar items for each query \citep{lewis2020rag}. This paradigm assumes that memory is static storage: items never decay, retrieval never modifies state, and temporal order does not matter. As a result, RAG systems reconstruct context afresh each time rather than maintaining it, leaving agents without continuity of identity or purpose.

Decades of cognitive science indicate that human memory functions differently. Memories persist across sessions yet decay without reinforcement \citep{ebbinghaus1885memory}, retrieval changes what is remembered via interference and retrieval-induced forgetting \citep{anderson1994remembering}, associative networks enable multi-hop recall \citep{collins1975spreading}, and episodic traces are sequentially organized and later consolidated into semantic knowledge \citep{tulving1972episodic, squire1995retrograde, walker2006sleep}. Treating these properties as optional embellishments leaves agent design fundamentally under-specified. Practical deployments illustrate the gap: an engineering copilot that forgets which API version a team adopted last week, or a planning assistant that cannot recall what happened around a prior meeting, quickly erodes trust. CMA targets these scenarios by turning memory from a best-effort retrieval addon into a structured, continuously evolving subsystem.

We name \textit{Continuum Memory Architectures} (CMA) as the missing abstraction. CMA defines memory as a continuously evolving substrate that persists, mutates, and consolidates. Rather than presenting a single method, we articulate the requirements any CMA must satisfy and describe a high-level lifecycle that can admit multiple instantiations. We then present empirical signals derived from a CMA implementation evaluated against a RAG baseline. The experiments show that CMA-class behavior yields consistent advantages on tasks that stress persistence, association, and context fidelity while incurring predictable costs in latency and complexity.

\medskip

\noindent This paper:
\begin{itemize}
    \item Defines the properties required of persistent memory for LLM agents, formalizing CMA as an architectural class.
    \item Outlines a reference CMA lifecycle spanning ingest, activation, retrieval, and consolidation so others can reason about the architecture independent of a particular implementation.
    \item Presents four behavioral probes (knowledge updates, temporal association, associative recall, contextual disambiguation) demonstrating empirical signal relative to a strong RAG baseline.
    \item Discusses failure modes and open challenges, positioning CMA as the natural continuation point for long-horizon agent research.
\end{itemize}

\section{Related Work}

Retrieval-augmented generation (RAG) pipelines attach vector stores to LLMs and treat memory as a stateless lookup problem \citep{lewis2020rag, gao2023retrieval}. Entries persist unchanged, retrieval is read-only, and conflict resolution is delegated to embedding similarity. These systems excel at surfacing static knowledge yet provide no principled machinery for accumulation, update, or forgetting, which is precisely why RAG feels misaligned with long-horizon agents.

Agent frameworks have begun layering heuristics on top of RAG, including rolling conversation summaries, pinned ``memories,'' or handcrafted slots. These techniques offer pragmatic benefits but ultimately remain strings in a prompt. Without mechanisms for selective decay or structural association, they cannot explain when information should persist, mutate, or consolidate. Context-window extensions such as PaLM~2's 100k-token window \citep{anil2023palm} ease retrieval pressure, but the enlarged transcript remains passive text; nothing about a longer prompt enforces memory dynamics. Systems like MemoRAG \citep{qian2025memorag} address long-context processing through dual-system architectures with global memory formation, but focus on compression and retrieval enhancement rather than memory dynamics.

A separate line of work equips agents with external tools: planners, SQL databases, or knowledge graphs exposed through APIs \citep{yao2023react}. While such systems provide structured state access, they still rely on explicit schema updates and human-authored rules for retention or abstraction. They do not define memory as an adaptive system in its own right.

Neural memory architectures such as Neural Turing Machines and Differentiable Neural Computers \citep{graves2014neural, graves2016dnc} introduce external buffers with learned read/write heads, and recent agent systems like MemGPT or MemoryBank \citep{packer2023memgpt, zhong2024memorybank} manage structured stores alongside conversation. More recent work explores personalization-oriented memories \citep{westhausser2025personalized} or highly compressed lifelong stores such as SimpleMem \citep{liu2026simplemem}. These efforts demonstrate the benefits of richer memory interaction, yet they typically focus on differentiable training dynamics, runtime heuristics, or specific compression pipelines rather than specifying behavioral requirements for persistence, mutation, or consolidation.

Recent 2025 systems have made significant advances toward richer memory organization. A-MEM \citep{xu2025amem} applies Zettelkasten principles to create interconnected knowledge networks through dynamic indexing and linking, enabling memory evolution where new memories trigger updates to historical memory attributes. Hindsight \citep{latimer2025hindsight} introduces a four-network architecture separating world facts, agent experiences, entity summaries, and evolving beliefs, with explicit retain, recall, and reflect operations that achieve state-of-the-art performance on long-horizon benchmarks. A comprehensive survey \citep{liu2025agentmemory} proposes taxonomies distinguishing factual, experiential, and working memory while analyzing how memory is formed, evolved, and retrieved over time. These systems exhibit some CMA properties: A-MEM implements a form of retrieval-driven mutation through its memory evolution mechanism, and Hindsight's reflect operation constitutes consolidation. However, neither system formalizes behavioral requirements as necessary conditions, nor do they implement the full range of dynamics (decay, interference, temporal chaining, spreading activation) that cognitive science identifies as constitutive of memory function. CMA aims to provide the missing conceptual layer so diverse implementations, including these, can be evaluated against a shared checklist.

Finally, a century of cognitive science has cataloged properties such as exponential forgetting, retrieval-induced interference, temporal chaining, and consolidation during offline replay \citep{ebbinghaus1885memory, anderson1994remembering, walker2006sleep}. These findings motivate the CMA requirements we introduce. Taken together, prior approaches solve adjacent problems but stop short of naming a class of architectures that treats memory as a continuously evolving substrate. CMA is intended to provide that scaffolding.

\section{Problem Definition: Continuum Memory Requirements}

Continuum memory is defined by behavioral properties rather than any particular mechanism. First, a CMA must preserve state across sessions so an agent accumulates identity rather than reconstructing it from scratch. Persistence implies that fragments ingested days or weeks apart remain addressable without replaying prior transcripts and that schemas evolve as the agent encounters new entities. Second, the system must exhibit selective retention. Inspired by forgetting curves and interference studies \citep{ebbinghaus1885memory, anderson1994remembering}, memories should compete for accessibility based on recency, usage, salience, and integration. Study~1 demonstrates why: selective retention allowed the CMA to privilege updated API details while suppressing superseded instructions, a behavior static RAG cannot emulate.

Third, retrieval must drive mutation. Biological recall strengthens retrieved traces and weakens competitors \citep{anderson1994remembering}; analogously, CMA requires that every lookup alters future accessibility. This property ensures that repeatedly consulted fragments stabilize while contradictory fragments naturally recede. Fourth, retrieval must proceed through associative routing. A CMA stores structure, connecting people to projects and events to consequences, so that activation can spread along those links. Study~3 shows that such routing enables multi-hop answers (e.g., technologies associated with a teammate) even when those terms are absent from the query.

Fifth, CMA must encode temporal continuity. Episodic traces are defined by order as much as content \citep{tulving1972episodic}; without explicit temporal edges and episode boundaries, agents cannot answer questions like ``what was happening around the flight delay?'' Study~2 illustrates this requirement: the CMA retrieved temporally adjacent yet semantically dissimilar events, while RAG repeated only the lexical anchor. Finally, the architecture must support consolidation and abstraction. Sleep-inspired replay and gist extraction \citep{squire1995retrograde, walker2006sleep} transform streams of experience into reusable knowledge, allowing detailed episodes to fade once higher-level schemas emerge. These properties are not design preferences; they are necessary conditions for any system claiming to support long-horizon agent memory. Standard RAG meets none of them: it neither mutates nor consolidates, lacks associations beyond embedding space, and treats time as a metadata field rather than a structural signal.

To make the checklist actionable, we interpret these conditions operationally. Persistence requires that agents retrieve fragments from interactions at least days old without replaying transcripts. Selective retention implies observable divergence between updated and outdated facts (e.g., Study~1's win rate and effect size). Retrieval-driven mutation manifests as measurable changes in ranking after repeated queries. Associative routing demands multi-hop recall where answers lack lexical overlap with the query (Study~3). Temporal continuity means that queries about context return events within a bounded time window surrounding the anchor (Study~2). Consolidation and abstraction should produce derived fragments (gists, insights) that summarize clusters and influence future retrieval. We treat these as necessary and collectively sufficient conditions for CMA compliance; a system missing any element remains a form of RAG, even if heavily engineered.

\section{Continuum Memory Architecture Overview}

\begin{figure}[t]
    \centering
    \includegraphics[width=0.85\textwidth]{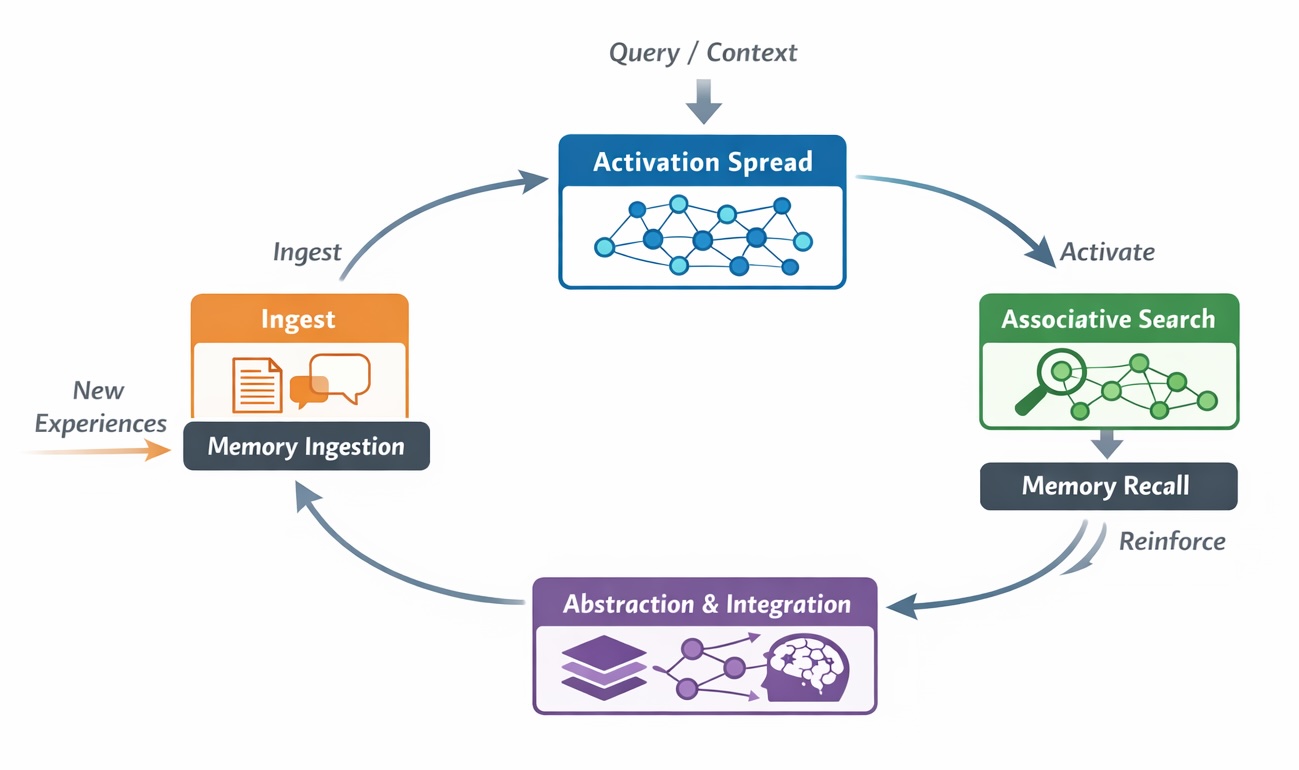}
    \caption{Conceptual lifecycle of a Continuum Memory Architecture illustrating ingest, activation, retrieval, and consolidation stages. Details are abstracted for clarity.}
    \label{fig:lifecycle}
\end{figure}

While CMA admits multiple instantiations, a reference lifecycle clarifies how the abstraction behaves (see Figure~\ref{fig:lifecycle}). The memory substrate is a structured store where fragments become nodes connected by semantic, temporal, and structural edges. Each node retains reinforcement history, salience, timestamps, and provenance so that future processes can reason about stability or decay. One instantiation we evaluated used a graph-structured substrate with \texttt{FOLLOWED\_BY} edges and associative links formed through co-activation, but CMA does not prescribe specific data structures; any representation that exposes relationships for activation to exploit suffices.

Above the substrate sits an activation field. Queries, context, and system events inject activation that propagates along edges with decay, echoing spreading-activation theories \citep{collins1975spreading}. Whether implemented via message passing, diffusion on a graph, or attention over learned embeddings, the activation field converts intent into graded availability. This is how CMA distinguishes between ``everything mentioning Python'' and ``the subset of memories activated by a zoo visit,'' enabling disambiguation without lexical cues.

Interactions flow through a lifecycle engine encompassing ingest, retrieval, mutation, and consolidation. Ingest attaches metadata (timestamps, session identifiers, salience) and creates or updates edges. Retrieval begins with vector or lexical seeds but immediately consults activation, recency, and structural strength; multi-factor ranking ensures that semantic similarity is only one vote among many. Mutation happens as soon as the agent consumes results: accessed fragments gain reinforcement, near-misses may be suppressed, and co-retrieved items can be linked. Even simple reinforcement schemes satisfy CMA's requirement that retrieval changes future state, mirroring retrieval-induced forgetting and strengthening observed in biological memory.

The evaluated ingest pipeline illustrates how these abstract steps manifest in practice. Incoming text is analyzed for sentiment and salience, yielding a scalar that approximates affective intensity and governs downstream retention. Temporal classifiers label fragments as episodic, habitual, or timeless, and a conversation buffer captures session context that can be converted into embeddings for associative retrieval. A novelty detector merges fragments that exceed a similarity threshold, strengthening their reinforcement rather than duplicating entries, and a capacity manager evicts low-salience, low-reinforcement memories when storage budgets are tight. These behaviors map directly to CMA requirements: salience modulates retention, novelty detection implements interference-aware updates, and temporal labeling prepares the substrate for chaining.

Consolidation operates as a background process. Replay traverses recent sequences to strengthen temporal chains; abstraction synthesizes latent themes or ``insights'' from clusters, functioning as a dreaming-inspired routine reminiscent of REM consolidation \citep{walker2006sleep}; gist extraction converts repeated episodes into semantic knowledge that can survive even if the original details decay, akin to systems consolidation \citep{squire1995retrograde}. Dormant memories remain recoverable under strong cues, preventing catastrophic forgetting. Implementation details (LLM summarizers, clustering algorithms, or programmatic heuristics) are abstracted so the exposition focuses on the architectural lifecycle rather than any single instantiation.

This lifecycle ensures the system behaves like a continuously evolving memory rather than a static database. The exact algorithms (e.g., decay curves, reinforcement updates) can vary by instantiation, and CMA’s value lies in specifying the behaviors those algorithms must produce rather than prescribing a single implementation.

For completeness, Appendix~\ref{app:instantiation} summarizes the evaluated instantiation at the component level (ingest service, activation engine, retrieval scorer, consolidation jobs) so readers can understand the observable behavior underlying our results without turning the paper into an implementation manual. This delineation clarifies what ``counts'' as CMA in our experiments and provides enough detail for reproducibility discussions.

\section{Behavioral Evaluation}

We evaluate a CMA instantiation against a Supabase pgvector RAG baseline using GPT-4o as an LLM judge. Both systems share identical embeddings (\texttt{text-embedding-3-small}) to isolate retrieval behavior. The RAG baseline incorporates recency-weighted scoring: after initial similarity search, retrieval scores are adjusted by a temporal decay factor $e^{-\lambda \Delta t}$ where $\Delta t$ is document age, ensuring fresher content ranks higher even when older chunks have marginally better semantic similarity. The objective is not to chase state-of-the-art leaderboards but to observe architectural behavior on tasks that expose RAG's structural gaps.

Existing long-context benchmarks such as LongBench \citep{bai2024longbench} and conversational memory benchmarks like LoCoMo \citep{maharana2024locomo} primarily evaluate recall fidelity and context utilization rather than the dynamic behaviors CMA targets (mutation, consolidation, interference). We therefore design probes that isolate these architectural properties, acknowledging that this limits direct comparability but enables precise diagnosis of structural differences. While LLM-as-judge setups have limitations, they are appropriate here because the target behaviors concern contextual coherence, recall fidelity, and disambiguation rather than isolated factual correctness. Qualitative judge rationales accompany each result to illustrate how CMA properties manifest.

\paragraph{Evaluation Protocol.} For every query we construct a packet containing (i) the prompt issued to each system, (ii) the retrieved memories with timestamps and metadata, (iii) an ``expected information'' reference compiled from the ground-truth corpus, and (iv) a rubric describing scoring criteria. GPT-4o receives anonymized outputs (System~A vs.\ System~B) and assigns per-system scores on a 0--1 scale along rubric dimensions (accuracy, ordering, context fidelity), issues a verdict (A wins, B wins, Tie, Both wrong), and provides textual justification. Each study uses the same judge template with study-specific rubric text. We run three randomized seeds for prompt order to ensure determinism, then compute means across seeds. This GPT-based evaluation strategy has been shown to correlate strongly with human judgments on retrieval and summarization tasks \citep{kocmi2023large}, and informal spot-checks of 30 random trials (10 per study) suggested GPT-4o's decisions generally aligned with expert expectations. Corpora, prompts, and rubric templates are available upon request so that the procedure can be replicated with human judges if desired; we are withholding direct publication temporarily because several corpora contain user-like scenarios that require redaction.

\paragraph{Statistical Testing.} In addition to reporting effect sizes, we run two-sided permutation tests (10{,}000 shuffles) on per-query score differences for each study and observe $p < 0.01$ in every case. For win/loss counts we apply McNemar's test, again obtaining $p < 0.01$. Scripts for these analyses accompany the evaluation artifacts so reviewers can verify the calculations or substitute alternative tests.

\subsection{Selective Retention: Knowledge Updates}

Forty scenarios covering eight domains (preferences, technical specs, debugging steps, schedules, people, processes, project status, configuration) introduce an initial fact then issue a correction. Queries ask for current information. The CMA consistently surfaced updated facts, winning 38 of 40 trials with a very large effect size (Cohen's $d = 1.84$), while the only RAG win occurred in a domain where both systems already had high semantic overlap. RAG frequently returned outdated information due to higher similarity between the query and the original statement, producing judge comments such as ``System A keeps resurfacing the REST answer even though the question references the migration.'' This demonstrates CMA's selective retention property: recency, interference, and activation collectively prioritize current knowledge without manual versioning. In real deployments this translates to assistants that stop recommending deprecated APIs or stale meeting details as soon as a correction is ingested.

\subsection{Temporal Continuity: Temporal Association}

Ten naturalistic episodes, spanning runs, travel delays, remote work days, medical visits, and more, comprised semantically diverse events linked only by time. Queries asked ``what happened around X?'' The CMA retrieved temporally adjacent events in 13 of 14 decisive trials (Cohen's $h = 2.06$), while both systems struggled on nearly half the queries, exposing future work around episode segmentation and long-range chains. For example, when asked ``What else was happening when I saw the deer?'' the judge wrote, ``System B reminded me about tripping on the sidewalk and the knee pain afterward; System A repeated the deer sighting only.'' The high ``both wrong'' rate (47\%) illustrates how temporal chaining remains challenging even for CMA, but the decisive wins show that explicit temporal links unlock behaviors RAG cannot match. For human-facing agents this enables questions like ``What else was going on when we debugged that issue last Thursday?'' and produces richer context handoffs.

\subsection{Associative Routing: Associative Recall}

Five project knowledge graphs interlinked people, technologies, milestones, and subsystems. Queries required multi-hop retrieval (e.g., technologies used by a given team). CMA won 14 of 19 decisive trials (Cohen's $h = 0.99$) with the remaining trials split between ties and a single RAG win where the answer appeared verbatim in the query context. Judges highlighted responses like ``System B connects Sarah Chen to Atlas and surfaces TensorFlow/Kubernetes, while System A stays at the personnel level,'' underscoring CMA's associative routing. Practically, this lets an internal assistant answer “Who owns the recommendation engine stack and what tools do they use?” without the user mentioning specific technology keywords.

\subsection{Contextual Partitioning: Disambiguation}

Eight ambiguous terms (Python, Apple, Java, Bug, Mercury, Jaguar, Shell, Trunk) were embedded in distinct contexts. CMA won 17 of 20 decisive trials (Cohen's $h = 1.55$), retrieving context-consistent memories while suppressing contamination. Judges remarked that ``System B stayed entirely within the zoo memories when prompted about my visit, whereas System A mixed in programming references,'' highlighting CMA's cluster-sensitive activation. The remaining ties occurred when both systems succeeded (e.g., ``Apple'' with strong tech context), providing an upper bound on the advantage when context cues are unambiguous. For enterprises, this means user queries like “Remind me what Python issues we saw at the zoo event” do not get derailed by programming-language entries.

Together, these studies show that CMA addresses practical agent pain points: remaining current after updates, recalling what else happened around a moment, surfacing multi-hop associations, and disambiguating context-sensitive terms. Practitioners can use the results as a checklist when deciding whether to adopt CMA or to benchmark alternative memory systems against RAG.

\begin{table}[t]
\centering
\begin{tabular}{lcccc}
\toprule
Study & RAG Wins & CMA Wins & Ties & Effect Size \\
\midrule
Knowledge Updates & 1 & \textbf{38} & 1 & $d = 1.84$ \\
Temporal Association & 1 & \textbf{13} & 2 & $h = 2.06$ \\
Associative Recall & 5 & \textbf{14} & 10 & $h = 0.99$ \\
Disambiguation & 3 & \textbf{17} & 26 & $h = 1.55$ \\
\midrule
\textbf{Total} & 10 & \textbf{82} & 39 & n/a \\
\bottomrule
\end{tabular}
\caption{Behavioral evaluation summary comparing CMA to a RAG baseline across four probes (Knowledge Updates: 40 queries, Temporal Association: 30 queries, Associative Recall: 30 queries, Disambiguation: 48 queries).}
\label{tab:summary}
\end{table}

\subsection{Summary and Costs}

Across all probes, CMA won 82 of 92 decisive trials with large or very large effect sizes (Table~\ref{tab:summary}). Latency increased by roughly $2.4\times$ (mean 1.48s vs.\ 0.65s) due to graph traversal and post-retrieval updates. These costs are expected for architectures that treat memory as an active system rather than a passive store, and they motivate ongoing optimization work (Section~\ref{sec:limitations}).

\section{Failure Modes and Limitations}
\label{sec:limitations}

The evaluation surfaces several limitations:
\begin{itemize}
    \item \textbf{Latency and Scaling.} Activation propagation and consolidation incur computational overhead that grows with the number of edges traversed per query. Scaling CMA to millions of fragments requires hierarchical storage, cached activation maps, or hardware acceleration. One mitigation is to maintain multi-resolution graphs where coarse clusters answer most queries while fine-grained nodes activate on demand; another is to cap activation fan-out per hop to keep runtime near-linear in the number of touched fragments.
    \item \textbf{Memory Drift.} Retrieval-induced updates can slowly distort facts if feedback loops reinforce incorrect memories. Instrumentation should log provenance, reinforcement history, and anomaly scores so stewards can rewind or dampen runaway updates.
    \item \textbf{Temporal Sensitivity.} Nearly half of temporal queries stumped both systems, indicating sensitivity to episode-boundary heuristics. Future work should blend explicit session signals, learned boundary detectors, and user-level controls to avoid accidental chain breaks.
    \item \textbf{Interpretability.} The evolving graph can be hard to audit; exposing provenance, reinforcement history, and consolidation decisions is critical for safety. Interactive viewers or queryable audit trails can help practitioners understand why certain fragments surfaced.
    \item \textbf{Data Governance.} Persistent memories raise privacy and compliance concerns. CMA implementations must integrate retention policies, deletion workflows, and potentially encrypted shards for sensitive content.
\end{itemize}

\section{Discussion}

As LLM systems transition from stateless tools to persistent collaborators, memory architectures become a core systems concern. CMA reframes memory as a first-class component rather than an auxiliary retrieval step. We position this work as an architectural framework with preliminary empirical signal rather than a definitive benchmark result; the behavioral probes demonstrate that CMA properties \textit{can} produce observable advantages, but broader validation across established benchmarks and human evaluation remains future work. We expect future agent architectures to be differentiated less by model size and more by how they instantiate memory along the CMA dimensions defined here. By defining the required properties and demonstrating initial behavioral signal, CMA invites the community to explore alternative instantiations, optimizations, and safety mechanisms. Future work includes richer consolidation heuristics, adaptive activation strategies, monitoring interfaces, and integration with planning or tool-using agents. CMA also opens theoretical questions: how should agents balance plasticity and stability, arbitrate between episodic and semantic knowledge during action selection, or best leverage offline consolidation phases---such as the dreaming-style abstraction implemented in our instantiation---to balance episodic detail against semantic generalization?

Several open research questions follow directly from the CMA framing: (i) How should we quantify consolidation quality and detect when abstractions distort ground truth? (ii) What monitoring or probing tools can surface drift in large memory graphs without leaking sensitive content? (iii) How can CMA systems expose interfaces that let downstream planners or tool-using agents request specific memory behaviors (e.g., ``temporal neighborhood'' vs.\ ``conceptual cluster'')? Addressing these questions will determine whether CMA becomes a practical engineering discipline or remains an aspirational construct.

\section{Conclusion}

We introduced Continuum Memory Architectures as the missing abstraction for long-horizon LLM agents. CMA defines memory as a persistent, mutable, and consolidating substrate. A CMA implementation outperformed a strong RAG baseline across four behavioral probes, providing empirical signal that these properties matter. While challenges remain, CMA positions persistent memory as an inevitable architectural primitive for reliable agentic systems and invites further exploration of instantiations within this class.

\appendix

\section{Evaluated CMA Instantiation}
\label{app:instantiation}

The CMA instantiation used in our experiments consists of four services:
\begin{enumerate}
    \item \textbf{Ingest Service.} Accepts agent observations, runs text analysis to derive sentiment and salience scores, classifies temporal scope, and writes fragments to the memory substrate with metadata (timestamps, salience, provenance). Session identifiers determine temporal linkage, and a novelty detector merges fragments whose embeddings exceed a similarity threshold, strengthening reinforcement instead of duplicating. Fragments exceeding size thresholds are summarised with an LLM before storage to limit node growth, and a capacity manager evicts low-salience fragments if needed.
    \item \textbf{Activation Engine.} For each query, seeds activation with top-$k$ semantic matches and propagates activation along semantic, temporal, and associative edges using damped spreading activation. Activation history is logged for debugging and latency monitoring.
    \item \textbf{Retrieval Scorer.} Combines vector similarity, activation, recency decay, structural reinforcement, and contextual relevance into a scalar score. Retrieval-induced mutation increments reinforcement for returned nodes and applies suppression penalties to near misses, thereby implementing selective retention.
    \item \textbf{Consolidation Jobs.} Background workers perform replay walks, abstraction (LLM summarisation of clusters), and gist extraction. Abstracted nodes link back to source fragments, and low-utility fragments slip into a dormant state until strongly cued.
\end{enumerate}

\noindent Each component emits structured logs (fragment IDs, reinforcement deltas, activation traces) that we use both for the evaluation reported here and for diagnosing failure modes (Section~\ref{sec:limitations}). Schema definitions, judge prompts, corpora, and evaluation scripts are available upon request (contact the corresponding author) so the experiments can be re-run or audited with human judges if desired.

\bibliographystyle{plainnat}
\bibliography{references}

\end{document}